\documentclass[letterpaper, 10 pt, conference]{ieeeconf} 
\IEEEoverridecommandlockouts
\usepackage{eqparbox}
\usepackage{multicol}
\usepackage{multirow}
\usepackage{diagbox}
\usepackage{slashbox}
\usepackage{amsmath}
\usepackage{algorithm}
\usepackage[noend]{algpseudocode}
\usepackage{array}
\usepackage{amssymb}
\usepackage{enumerate}
\usepackage{graphicx}
\usepackage{amsmath}
\usepackage{epstopdf}
\usepackage{url}
\usepackage{booktabs}
\usepackage{breqn}
\usepackage{afterpage}
\usepackage{cite}

\usepackage{caption}
\usepackage{tikz}
\usetikzlibrary{arrows,calc,positioning}
\tikzstyle{intt}=[draw,text centered,minimum size=6em,text width=5.25cm,text height=0.34cm]
\tikzstyle{intl}=[draw,text centered,minimum size=2em,text width=2.75cm,text height=0.34cm]
\tikzstyle{int}=[draw,minimum size=2.5em,text centered,text width=3.5cm]
\tikzstyle{intg}=[draw,minimum size=3em,text centered,text width=6.cm]
\tikzstyle{sum}=[draw,shape=circle,inner sep=2pt,text centered,node distance=3.5cm]
\tikzstyle{summ}=[drawshape=circle,inner sep=4pt,text centered,node distance=3.cm]

\begin{document}

\title{Driving Intention Recognition and Lane Change Prediction on the Highway}

\author{Teawon~Han, ~Junbo~Jing,~and~{\"U}mit~{\"O}zg{\"u}ner 
\thanks{Teawon Han, Junbo Jing, and {\"U}mit {\"O}zg{\"u}ner are with Department of Electrical and Computer Engineering, The Ohio State University, Columbus, Ohio, USA
        {\tt\small  \{han.394, jing.71, ozguner.1\}@osu.edu }}%
}

\maketitle

\begin{abstract}
This paper proposes a framework to recognize driving intentions and to predict driving behaviors of lane changing on the highway by using externally sensable traffic data from the host-vehicle. The framework consists of a driving characteristic estimator and a driving behavior predictor. A driver's implicit driving characteristic information is uniquely determined and detected by proposed the online-estimator. Neural-network based behavior predictor is developed and validated by testing with the real naturalistic traffic data from Next Generation Simulation (NGSIM), which demonstrates the effectiveness in identifying the driving characteristics and transforming into accurate behavior prediction in real-world traffic situations. 
\end{abstract}

\IEEEpeerreviewmaketitle

\section{Introduction}

The prediction of other vehicles' future behavior is required for automated cars to make a right decision under the mixed traffic conditions where human-driven and automated cars are on the same road. In this work, our goal is the recognition of driving intention and the prediction of driving behaviors on the highway.


Different methodologies have been proposed to predict driving behaviors on the intersection\cite{kurt2010hybrid,gadepally2014framework, amsalu2015driver, kuge2000driver}, ramp\cite{meng2012classification,wei2013autonomous,dong2017intention}, and highway\cite{wissing2017lane, kumar2013learning, dou2016lane}.
Transition of driving behavior on the intersection is represented by FSM and the future driving behavior is predicted in \cite{kurt2010hybrid}. In \cite{gadepally2014framework, kuge2000driver}, Hidden Markov Models (HMM) are trained for individual driving behaviors, and used to predict future behaviors, while Support Vector Machines (SVM) is proposed in \cite{amsalu2015driver}.
For merging behaviors, a classification and regression tree (CART) is created and trained in \cite{meng2012classification}. For highway ramp entrances, \cite{wei2013autonomous} implements Bayes' theorem to obtain the probability of driving intentions (yield or not), and \cite{dong2017intention} proposes a probabilistic graphical model. For highway lane-change behavior predictions, situation and movement based features are implemented separately in \cite{wissing2017lane} to calculate individual probability of lane-change behavior, and are combined and considered as a final prediction. Multiclass classification via SVM and Bayesian filtering are used to recognize lane-change intentions in \cite{kumar2013learning}. SVM and Artificial Neural Network (ANN) classifiers are combined and proposed to predict lane-change at highway lane drops in \cite{dou2016lane}. Under HSS framework, HMM based driving behavior models are dynamically grafted and pruned concerning situations to increase prediction accuracy in \cite{gadepally2017framework}. Instead of detecting maneuver-patterns of driving behaviors, determination and classification of driving style has been investigated in \cite{ma2007behavior,derbel2015driving,liu2014trajectory,liu2015classification}. 

In previous studies, behavior prediction models are designed to detect maneuver-patterns of driving behaviors from collected traffic data. However, the pattern of an identical behavior is various depending on traffic conditions. Also, it is practically unfeasible to collect a large enough amount of data including patterns under all possible traffic conditions. Therefore, the performance of prediction model which is seeking maneuver-patterns depends on the data. In this study, it is focused on determination and detection of general driving attributions which are independent of the traffic conditions, and its transition patterns are detected and implemented to train and test the behavior prediction model.

\begin{figure}[t!]
\centering 
\includegraphics[width=7.2cm]{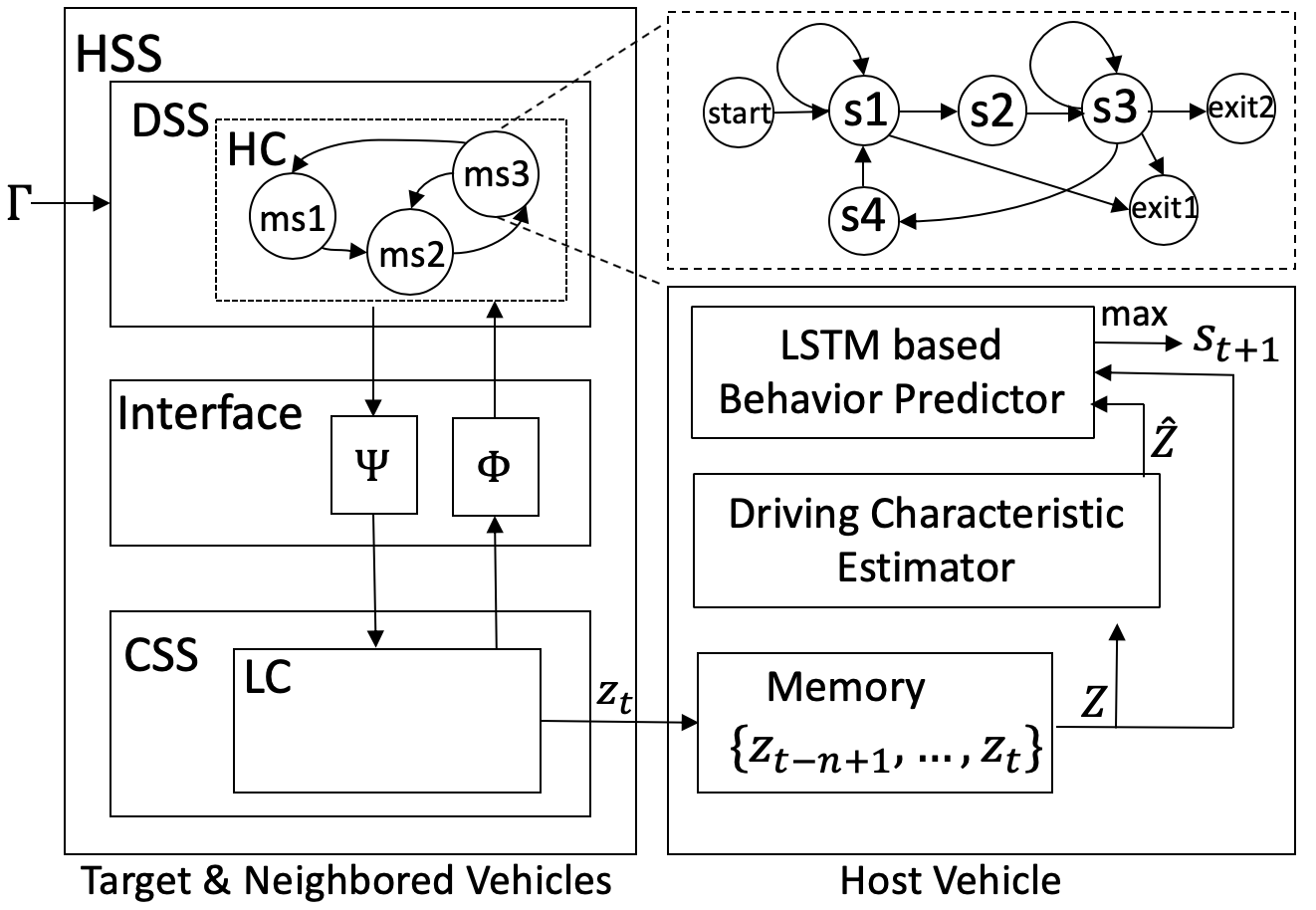}
\caption{The proposed system framework for driving intention recognition and driving behavior prediction}
\label{fig:overview_system}
\vspace{-0.8cm}
\end{figure}

In the framework as shown in figure \ref{fig:overview_system}, the control system structure of target and neighbour vehicles are represented by the Hybrid State System (HSS) which was first proposed to control the automated car (OSU-ACT) under various traffic situations in the 2007 DARPA Urban Challenge as shown in \cite{kurt2008hybrid, redmill2008ohio}, and has been implemented to represent the control structure in \cite{kurt2013hierarchical, gadepally2017framework, liu2014trajectory,liu2015classification}.
The proposed methodology consists of a driving characteristic estimator and a driving behavior predictor. The uniquely determined driving-characteristics consist hidden general longitudinal and lateral driving attributions, of which the sequential transitions are considered as the driving intentions. 
After estimating driving characteristics, general transition patterns of the estimations are recognized by training the neural network based behavior predictor. Three driving behaviors of Lane Change Left (LCL), Lane Change Right (LCR), and Lane Keeping (LK) on the highway are considered, and the future driving behavior is obtained by choosing a behavior which has the highest probability in prediction.
In this paper, Section \ref{section:Detection of driving characteristics through deterministic driver and lane-change models} introduces the proposed system framework, and defines the driving characteristics by using the Intelligent Driver Model (IDM) and the Minimize Overall Braking Induced by Lane change (MOBIL) model.
Section \ref{section:Estimation methods for longitudinal driving characteristics} describes the parameter estimation method for driving characteristics using practically sensable inter-vehicle state data with IDM and MOBIL models. Section \ref{section:Driving Behavior Predictor} describes the structure of the behavior predictor by a LSTM network, and illustrates the accuracy of prediction results. Finally, the contributions are concluded in Section \ref{section:conclusion}.     

\section{System framework and determination of driving characteristics}\label{section:Detection of driving characteristics through deterministic driver and lane-change models}


\subsection{Proposition of System framework }\label{subsection:Proposition of System framework}
The structure of control system for target and neighbour vehicles is represented by HSS in figure \ref{fig:overview_system} because HSS is able to clearly represent the relation between driver and vehicle states as shown in \cite{kurt2008hybrid, redmill2008ohio, kurt2013hierarchical}. 
Because internal transitions of other vehicles are not detectable without communication, the proposed framework is outside of HSS, and the system input is only externally sensable data: vehicle-positions. The recent few seconds sensing data is stored and used for estimation of the driving characteristics and prediction of driving behaviors. The behavior predictor returns probabilities of LCL, LCR, and LK. Note that the vehicle-state consists of longitudinal position, velocity, and lane-number, notating the state of vehicle $\alpha$ by $w_t(\alpha)=\left[pos_t\left(\alpha\right), vel_t\left(\alpha\right), lane_t\left(\alpha\right)\right]$. The sensing data $z_t$ includes $m$ number of vehicle-states at time-step $t$ expressing by $z_t=\left[w_t(veh_1),w_t(veh_2), ..., w_t(veh_m)\right]^T$. In figure \ref{fig:overview_system}, the stored data  $Z=\left\{z_{t-n+1},...,z_{t-1}, z_{t}\right\}$ is used to estimate driving characteristics $\hat{Z}=\left\{\hat{z}_{t-n+1},...,\hat{z}_{t-1}, \hat{z}_{t}\right\}$ where $\hat{z}_{t}=\left[c^1_t, c^2_t, ..., c^j_t\right]^T, j=\left\{1,...,k\right\}$; $c^j_{t}$ is an estimated $j^{th}$ driving characteristic at $t$, $n$ is size of look-back horizon, and $k$ is the number of driving characteristic parameters. Both, $Z$ and $\hat{Z}$, are transferred to the LSTM based behavior predictor to obtain probabilities of target-vehicle's future driving behavior $s_{t+1} \in \left\{LCL, LCR, LK\right\}$.
\subsection{Intelligent Driver Model(IDM) and Minimize Overall Braking Induced by Lane change (MOBIL)} \label{subsection:Intelligent Driver Model(IDM) and Minimize Overall Braking Induced by Lane change (MOBIL)}
\begin{figure}[h!]
  \centering
  \includegraphics[width=7cm]{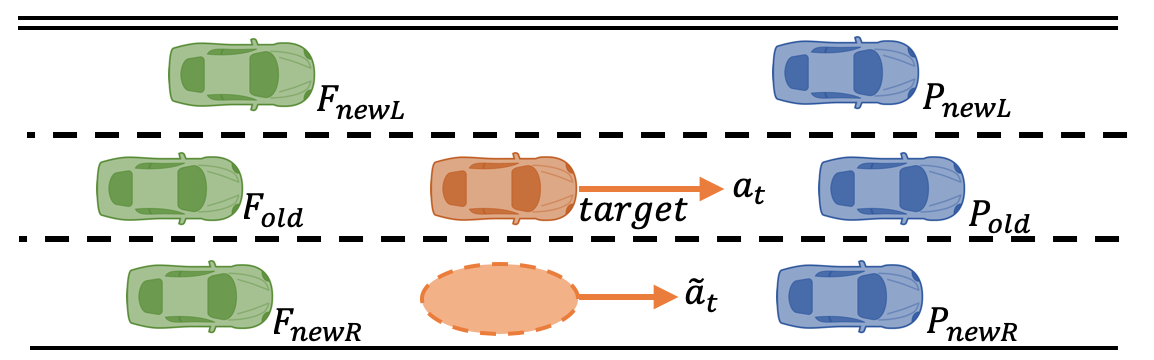}
  \caption{Annotation of nearby vehicles for MOBIL Model}
  \label{fig:determination_of_followingveh}
\end{figure}

 The IDM \cite{treiber2000congested}\cite{kesting2010enhanced} is a crash-free microscopic car-following model which estimates longitudinal dynamic speed controls. It is configured by several parameters which reflect potential factors of the following vehicle's driving behavior as shown in equation (\ref{eq:IDM1}), (\ref{eq:IDM2}), and (\ref{eq:IDM3}).
\begin{equation}\label{eq:IDM1}
\dot{x_{k}}= \frac{dx_{k}}{dt}={v_{k}}
\end{equation}
\begin{equation}\label{eq:IDM2}
\dot{v_{k}}=\frac{dv_{k}}{dt}=\alpha \left ( 1-\left ( \frac{v_{k}}{v_{0}} \right )^\delta -\left ( \frac{s^{*}\left (v_{k},\Delta v_{k}\right)}{s_{k}}\right )^2 \right )
\end{equation}
\begin{equation}\label{eq:IDM3}
\begin{split}
\begin{aligned}
\text{where }
s^{*}\left (v_{k},\Delta v_{k}\right) & =s_{0}+v_{k}T+\frac{v_{k}\Delta v_{k}}{2\sqrt{\alpha \beta}} \\
\Delta v_{k} & =v_{k} - v_{preceding}
\end{aligned}
\end{split}
\end{equation}
 $\protect v_{k}$ and $\protect \dot{v_{k}}$ are velocity and acceleration of following vehicle $k$. $s_{k}$, $s_{0}$, and $\Delta v_{k}$ refer gap, desired gap, and  velocity difference between preceding  and following vehicles respectively. $\alpha, \beta, \delta$, and $T$ are desired acceleration, desired deceleration, IDM order coefficient, and desired safety time headway respectively. The MOBIL \cite{treiber2009modeling,kesting2007general} is a lane-change decision-making model which automatically derives lane-change policy based on nearby longitudinal traffic conditions as shown in equation \ref{eq:MOBIL1} and \ref{eq:MOBIL2}.
\begin{equation}\label{eq:MOBIL1}
\tilde{a}_t^{N}\geq -b_{safe}
\end{equation}
\begin{equation}\label{eq:MOBIL2}
\tilde{a}_t^{T}-{a}_t^{T}+p\left( \tilde{a}_t^{N}-{a}_t^{N}+\tilde{a}_t^{O}-{a}_t^{O}\right)> \Delta a_{th}
\end{equation}
$a_t^{T}$, $a_t^{N}$, and $a_t^{O}$ are expected accelerations of target, new following, and old following vehicles respectively at time-step $t$ without lane-change; $\tilde{a}_t^{T}$, $\tilde{a}_t^{N}$, and $\tilde{a}_t^{O}$ are expected accelerations with lane-change at time-step $t$. Old and new following vehicles are determined based on target-vehicle and direction of lane-change as shown in figure \ref{fig:determination_of_followingveh}. $P_{old}$ and $F_{old}$ are the preceding and the following vehicles before the target-vehicle changes a lane. $F_{newL} \left(P_{newL}\right)$ and $F_{newR}\left(P_{newR}\right)$ are vehicles that will follow (precede) the target-vehicle after changing a lane to the left and right respectively. $p$ is a politeness factor that determines how much others advantages will be considered for the lane-change decision-making. The lane-change behavior is selected when left-hand side of equation \ref{eq:MOBIL2} is greater than a threshold, $\Delta a_{th}$.  
\vspace{-3.8mm}
\subsection{Definition of driving characteristics}
\label{subsection:Determination of driving characteristics}
The IDM parameters don't vary if driving propensity is identical under various traffic situations. Therefore, it is possible to extract general longitudinal driving characteristics from traffic data by estimating parameters. MOBIL is a cooperative lane-change decision making model. It is chosen to obtain lateral driving characteristics because it can be used to quantify conditions that drivers change a lane. For quantification, equation \ref{eq:MOBIL2} is modified to equation \ref{eq:incentive} as described in Section \ref{section:Estimation of lateral driving characteristics}, so that the incentives of LCL and LCR are obtainable at every time-step. In this study, two IDM parameters, $a_{t}$ and $T_{t}$, and two incentives, $I_{t}^{lcl}$ and $I_{t}^{lcr}$, are considered as longitudinal and lateral driving characteristics. The recent past three-seconds data with 0.1 seconds unit-time is stored into the memory for estimation of the driving characteristics. Therefore, the general form of $\hat{Z}$ defined in Section II-A can be specified such as $\hat{Z}=\left\{\hat{z}_{t-30+1},...,\hat{z}_{t-1}, \hat{z}_{t}\right\}$ where $\hat{z}_{t}=\left[T_{t}, a_{t}, I^{lcl}_{t}, I^{lcr}_{t}\right]^T$. The details of estimation-setting and process are described in the section \ref{section:Estimation methods for longitudinal driving characteristics} and \ref{section:Estimation of lateral driving characteristics}.
\section{Estimation methods for longitudinal driving characteristics}
\label{section:Estimation methods for longitudinal driving characteristics}

\subsection{Optimization based driver parameter estimation} \label{subsection:optimization based driver parameter estimation}
The core of the handled estimation problem is in fact an optimization problem, of which the optimization variables are the selected IDM parameters, and the cost function is a defined fitting error between the modeled car-following response and the measured one:
\begin{equation}\label{eq:Costfunc}
\min J = \left| {{f^{IDM}}\left( {\vec x}, {w_t(v_T)}, {w_t(v_P)} \right) - \vec y} \right|
\end{equation}
where ${{\vec x}}$ refer to the set of selected IDM parameters, ${\vec y}$ is the estimation target vehicle's measured car-following response state vector, and ${{f^{IDM}}}$ refers to the IDM model functions as described in equation (\ref{eq:IDM1}), (\ref{eq:IDM2}), and (\ref{eq:IDM3}) with the states of target and preceding vehicles at $t$, $w_t(v_T)$ and $w_t(v_P)$.

Though the IDM model contains six parameters in total, as has been discovered in \cite{punzo2015we}, it is not necessary to estimate all six of the parameters due to the parameters' sensitivity influence to the model's response. Also, the amount of estimated parameters / optimization variables further complicates the optimization problem and makes it difficult to converge. In \cite{punzo2015we}, a variance based sensitivity analysis is performed on the IDM using NGSIM data, which is the same data source as in this study, and it is discovered that only three IDM parameters are of critical influence when approximating real-world driving motion with IDM. The three most influential parameters are desired safety time headway ${T}$, IDM order coefficient ${\delta }$, and desired acceleration ${a}$. Furthermore, the sensitivity significance of desired time headway ${T}$ is significantly larger than the other two. For detailed sensitivity quantification, refer to \cite{punzo2015we}.

For online estimation of driver parameters, the problem's input-output responses are measured through a moving horizon of the recent time, which includes the speed traces of the preceding vehicle and the follower vehicle, and the inter-vehicle distance trace. Vehicle acceleration, as being one state of IDM in equation (\ref{eq:IDM2}), is generated by taking forward derivative of speed profile in discrete time. The objective of fitting is defined by the modeled follower vehicle's acceleration response truthfulness to the preceding vehicle.  

To maintain physical meaning of the IDM parameters, the optimization process of minimizing fitting error should be bounded within a reasonable parameter value range.
Further, the IDM model's response stability is maintained only in a valid range of the parameters and the state variables. Therefore, the three optimization variables are hard bounded, and the range is selected with reference to \cite{treiber2000congested}\cite{punzo2015we}, and is decided at $\protect \delta \in \left [ 3.8, 4.2 \right ]$, $\protect T \in \left[0.1, 5.0 \right] \left(s\right)$, and $\protect a \in \left[0.1, 9.0 \right] \left(m/s^2\right)$.


\subsection{Optimization by Genetic Algorithm with Guided Initialization by eTS Data Online Clustering}
The optimization problem formulated by the functions of IDM imposes high nonlinearity and high order, due to the locations of optimization variables in the function and the recurrent way of mode state update in discrete time. For such type of problem, a commonly used capable optimization method is by Genetic Algorithm (GA)\cite{houck1995genetic}.
In this study, the Matlab GA toolbox is applied to solve the formulated problem.
The evolution Takago-Sugeno online clustering method \cite{angelov2004approach,filev2011real} is implemented to determine proper initial population. It helps optimization methods to get a better solution as shown in \cite{jing2015vehicle}. In a similar way, selected three IDM parameters are estimated by using GA with eTS online clustering. The estimated IDM parameters at time-step $t$ are denoted by $\theta_{t}^{*} = \left[ \delta_{t}^{*}, T_{t}^{*}, a_{t}^{*} \right]^T$. In every estimation step, the constraints of three variables in the optimization problem are updated, and the clustering method guides this process.
The eTS online clustering creates a new cluster or updates existing clusters dynamically based on a new estimation $\theta_{t}^{*}$, following the specific steps described in algorithm \ref{alg:etsalgorithm}. The centers of existing clusters are kept and expressed as $C_t=\left\{c_t^1, c_t^2, ..., c_t^i \right\}$, where $i$ is the index of clusters at time-step $t$, $c_t^i \in R^3$, and the total number of clusters is not initially fixed. After clustering, a center of new cluster or one of existing cluster centers which has the highest similarity to $\theta_{t}^{*}$ is selected. New constraints are created by assigning the selected cluster center $c_t^{*}$ to the mean of constraint-range as shown in equation (\ref{eq:updateconstraints}) where $\gamma_{1}$ and $\gamma_{2}$ are scale factors which are set by 0.55 and 1.45 in this study; $c_t^{*} = \{\theta_{t}^{*}$ or $c_t^l\}$ in algorithm \ref{alg:etsalgorithm}.

\begin{algorithm}
\caption{On-line clustering based on eTS method}\label{alg:etsalgorithm}
\begin{algorithmic}[1]
\Procedure{ets clustering func }{$\theta_{t}^{*}, Z_{t-1}, C_{t}$}
\State $n \gets$ size$\left(C_{t}\right)$
\State $i=\left\{ 1, ..., n \right\}$
\State $Z_t=\left\{Z_{t-1},\theta_{t}^{*}\right\}$ where, $Z_{t-1}=\left\{z_1, ..., z_{t-1}\right\}$
\State $P_{t}\left( z_{t} \right) \gets$ Calculate the potential of $\theta_{t}^{*}$ by Eq. $\left(\ref{eq:etspotential}\right)$
\State $P_{t}\left( c_{t}^{i} \right) \gets$Update the potential of $c_{t}^{i}$  $\forall i$ by Eq. $\left(\ref{eq:etsupdate}\right)$
\If {$P_{t}\left( \theta_{t}^{*} \right) > max\left[ P_{t}\left( c_{t}^{i}\right) \forall i \right]$}
\State $s=\underset{i}{argmin}\left\| \theta_{t}^{*} - c_{t}^{i}\right\| $ $\forall i$
\If {$ \left\| \theta_{t}^{*} - c_{t}^{s} \right\|<\epsilon $}
\State $c_{t}^{s}:=\theta_{t}^{*}$, where $c_{t}^s \in C_{t}$
\Else
\State  Insert $\theta_{t}^{*}$ into $C_{t}$  
\EndIf
\State \textbf{return} $Z_t$, $C_{t}$ and $\theta_{t}^{*}$
\Else
\State $\lambda_{t}^{i} \gets$ Calculate similarity $\left(\theta_{t}^{*}, c_{t}^{i}\right)$, $\forall i$ by Eq. (\ref{eq:etssimilarity})
\State $l=\underset{i}{argmax}\left( \lambda_{t}^{i}\right) \forall i$
\State \textbf{return} $Z_t$, $C_{t}$ and $c_{t}^{l}$
\EndIf
\label{euclidendwhile}
\EndProcedure
\end{algorithmic}
\end{algorithm}\vspace{-0.5cm}

\begin{equation}\label{eq:updateconstraints}
\begin{split}
\delta_{t+1} &\in \left [ c_{t}^{*}\left(1\right)\times \gamma_{1}, c_{t}^{*}\left(1\right)\times \gamma_{2}\right ] \\
 T_{t+1} &\in \left[c_{t}^{*}\left(2\right)\times \gamma_{1}, c_{t}^{*}\left(2\right)\times \gamma_{2} \right]\left(s\right)\\
 a_{t+1} &\in \left[c_{t}^{*}\left(3\right)\times \gamma_{1}, c_{t}^{*}\left(3\right)\times \gamma_{2} \right]\left(m/s^2\right)
\end{split}
\end{equation}

\begin{equation}\label{eq:etspotential}
P_{t}\left( z_{t}\right)=\frac{t-1}{(t-1)(a_{t}+1)-2c_{t}+b_{t}}
\end{equation}
where
\begin{equation}
a_{t}=z_{t}^{T}z_{t}
\end{equation}
\begin{equation}
b_{t}=b_{t-1}+z_{t-1}^{T}z_{t-1}
\end{equation}
\begin{equation}
c_{t}=\sum_{j=1}^{3}\left( z_{t}^{j}\sum_{k=1}^{t-1}z_{k}^{j}\right)
\end{equation}
\begin{equation}\label{eq:etsupdate}
P_{t}(c_{t}^{i})=\frac{(t-1)P_{t-1}(c_{t}^{i})}{t-2+P_{t-1}(c_{t}^{i})[1+q\left\|c_{t}^{i}-z_{t-1}\right\|^2]}
\end{equation}
\begin{equation}\label{eq:etssimilarity}
\lambda_{t}^{i}(z_{t})=\frac{\gamma_{t}^{i}(z_{t})}{\sum_{j}\gamma_{t}^{j}(z_{t})}
\end{equation} 
\begin{equation}\text{where    }  
\gamma_{t}^{i}(z_{t})=exp\left(-\frac{(z_{t}-c_{t}^{i})^{T}(z_{t}-c_{t}^{i})}{var(c_{t}^{i})}\right)
\end{equation}
$c_{t}^{i}$ is a $i^{th}$ cluster center at time-step $t$. A coefficient $q$ in equation (\ref{eq:etsupdate}) and a constant threshold value $\epsilon$ in algorithm $\ref{alg:etsalgorithm}$ affect to the decision making of creating new clusters. In this study, $q$ and $\epsilon$ are set by $7$ and $0.45$ respectively. The output of algorithm \ref{alg:etsalgorithm}, $\theta_{t}^{*}$ or $c_t^l$, is implemented to update constraints of IDM parameters for next estimation process as described above. 

\begin{figure}
\centering
\includegraphics[width=7cm]{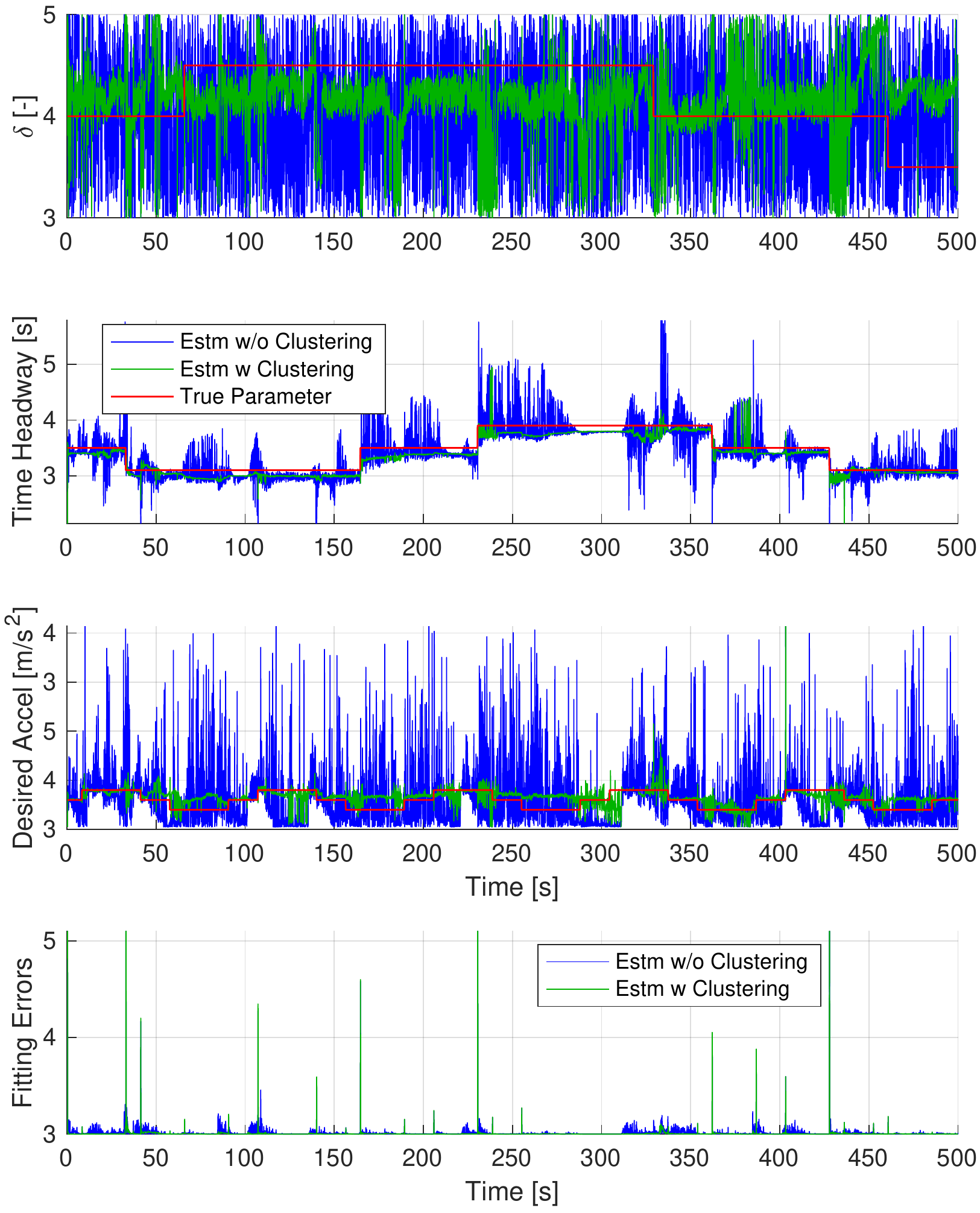}
\caption{Comparison of IDM parameter estimation by GA with and without clustering in simulated traffic data}
\label{fig:Sim_Estm_Compare}
\end{figure}

To evaluate performance of the IDM parameter estimation, car-following is simulated by using IDM with arbitrary set parameters. Since the data is noiseless by generating from pure IDM model, the accuracy benchmark is easily set at zero fitting error. As shown by Figure \ref{fig:Sim_Estm_Compare}, variations on the IDM parameters are created using square waves. Because the estimation problem is formulated using a single set of parameters to approximate the response of a past horizon in GA method, the fitting error spikes at the parameter transition points can be observed as the model of a single parameter set cannot fit the response by transiting two sets of parameters. For other cases, the fitting error is near zero. Algorithm-wise, the estimator with clustering clearly outperforms the one without clustering in fitting error, and with MAE at 0.0015 and 0.0032 respectively.
Also, the estimated parameters are noticeably more calm when with data online clustering. Especially the result of estimated time headway parameter very well recreated the true simulated parameter values. Result chattering can be observed on the other two parameters. This is because the sensitivities of the IDM's order parameter and the desired acceleration parameter are significantly less than the desired time headway parameter, as has been discovered in \cite{punzo2015we}. Multiple convergences exist for minimal fitting error, and the two less sensitive parameters can compensate each in generating a good local fitting. 

Using NGSIM US I-80 highway real traffic data \cite{NGSIM2008a}, the IDM parameters are estimated through proposed optimization methods, GA with eTS based online clustering. Due to the lack of ground truth knowledge of IDM parameters in the real traffic data, estimation performance is shown through the fitting errors. The 238, 80, and 1630 number of driving samples are extracted from the NGSIM data for LCL, LCR, and LK respectively. Among total observation horizon of the samples, three seconds driving data before the transition of driving-behavior is chosen to analyze the performance of IDM parameter estimation. 
This is because that the last few seconds before the transition of driving behavior is the period that driver's propensity is varying, which makes difficult to estimate parameters. 
The unit time of NGSIM data is $0.1$ seconds, therefore, the 30 number of estimations are executed for three seconds in every driving sample. The GA optimization method results in low fitting errors as shown in table $\ref{tab:fittingerror}$. 
\begin{table}[h]
\caption{IDM Parameter Fitting Errors $\left[ \% \right]$}
\centering
\resizebox{.83\columnwidth}{!}{
\begin{tabular}{|c||*{3}{c|}}\hline
\backslashbox{Errors}{Cases}
&\makebox[3em]{LCL}&\makebox[3em]{LCR}&\makebox[3em]{LK}\\\hline\hline
$E<0.1$ &93.17&91.21&97.14\\\hline
$0.1 \leq E <0.3$ &1.15&1.25&0.56\\\hline
$0.3 \leq E <0.5$ &0.43&0.54&0.26\\\hline
$0.5 \leq E$ &5.25&7.0&2.04\\\hline\hline
\# of samples &238&80&1630\\\hline
\end{tabular}
}\vspace{-3mm}

\label{tab:fittingerror}
\end{table}
\section{Estimation of lateral driving characteristic}
\label{section:Estimation of lateral driving characteristics}
As described in Section \ref{section:Detection of driving characteristics through deterministic driver and lane-change models}, it is proposed to implement MOBIL, the lane-change decision-making model, to obtain the lateral driving characteristics. A key idea of MOBIL is the incentive calculation for lane-change decision-making as shown in the equation (\ref{eq:MOBIL2}). Left-hand side of the equation returns expected incentive obtained by considering not only target-vehicle, but also neighbour following vehicles. 

The politeness factor $p$ defines how much other vehicles' benefits will be concerned in calculating the incentive. In \cite{treiber2009modeling,kesting2007general}, the range of general politeness-factor is defined as $0.2<p<0.5$ through analysis of real lane-change traffic data. 
The politeness factor is treated as a tunable parameter for a region's normal traffic situation in current work, and is selected at the constant value $p=0.35$. The parameter will be made self-adaptive for a given traffic data set in the future. Following the metric for inequality condition in (\ref{eq:MOBIL2}), the incentives for lane-change left $I_{t}^{lcl}$ or lane-change right $I_{t}^{lcr}$ maneuver at time-step $t$ can be quantified as in equation (\ref{eq:incentive_concised}), (\ref{eq:incentive}), which are also treated as lateral driving characteristics.
\begin{equation}\label{eq:incentive_concised}
I_{t}^{d}=f^{Inc}\left(\tilde{a}_{t, \left(d\right)}^{T}, {a}_{t}^{T}, \tilde{a}_{t}^{N}, {a}_{t}^{N}, \tilde{a}_{t}^{O}, {a}_{t}^{O}\right), d=\left\{lcl, lcr\right\} 
\end{equation}
\begin{equation}\label{eq:incentive}
f^{Inc}=\tilde{a}_{t, \left(d\right)}^{T}-{a}_{t}^{T}+0.35\left( \tilde{a}_{t}^{N}-{a}_{t}^{N}+\tilde{a}_{t}^{O}-{a}_{t}^{O}\right) 
\end{equation}

\section{Driving Behavior Predictor}\label{section:Driving Behavior Predictor}


\subsection{Structure of driving behavior predictor}
\begin{figure}[h!]
\centering
\includegraphics[width=7.5cm]{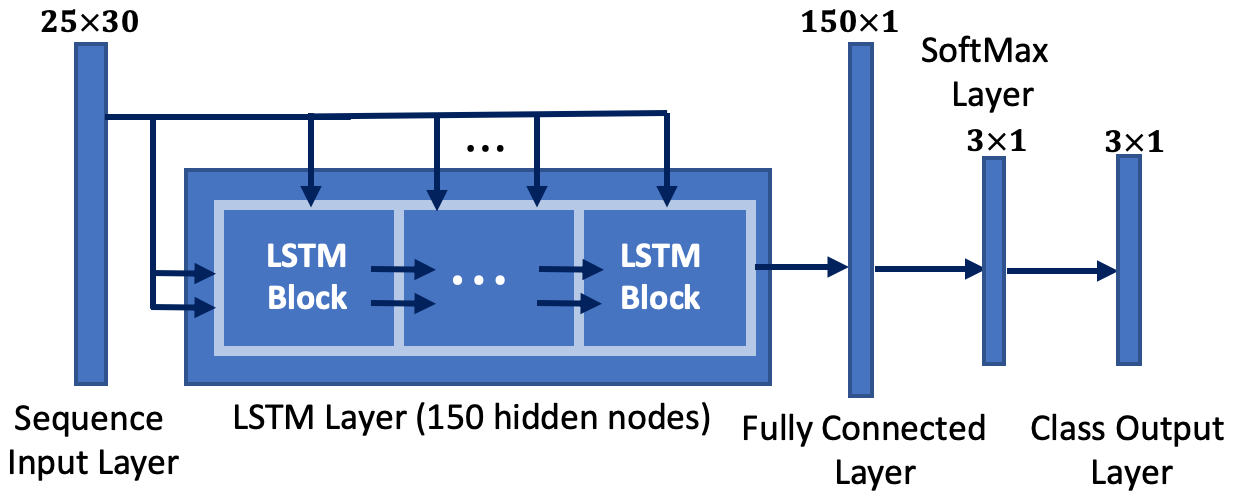}
\caption{Structure of the driving behavior predictor.}
\vspace{-0.3cm}
\label{fig:LSTM_predictor}
\end{figure}

Unlike RNN, Long Short-Term Memory (LSTM) network \cite{hochreiter1997long} can detect dependencies among long-horizon data by storing relevant information into memory cells. Thus, a LSTM network is implemented to detect general patterns and classify them into LCL, LCR, and LK in three seconds sequential sensing data and estimated driving characteristics. 
The driving behavior predictor consists of five neural-network layers as shown in figure \ref{fig:LSTM_predictor}. At every time-step, three seconds of sensing data $Z_t$ and driving characteristics $\hat{Z}_t$ are fed to the sequence input layer. As the activation function, a sigmoid function $\sigma(x)=(1+e^{-x})^{-1}$ is implemented in the network.
The size of hidden states is set by 150 in the LSTM layer. Classification results are the outcomes from the fully-connected layers. A softmax layer is inserted to make output range to $\left[0, 1\right]$. In the classification output layer, a cross-entropy function is applied as the loss-function ($\ref{eq:lossfunction}$) for multiclass classification \cite{bishop2006pattern}, where $y_i$ and $\hat{y}_i$ are the binary indicator (0 or 1) and the prediction of $i^{th}$ class respectively. Classes are defined by LCL, LCR, and LK manuevers. MATLAB deep learning toolbox is applied to design and to train the proposed neural network. 
\begin{equation}
\label{eq:lossfunction}
loss=-\sum_{i=1}^{3}y_{i} ln {\hat{y}_{i}}
\vspace{-0.3cm}
\end{equation}
\subsection{Training data and process}

The proposed driving behavior predictor network is trained by using the naturalistic traffic data and the estimated driving characteristics. Specifically, traffic data which can be feasibly sensed by the target-vehicle is generated from NGSIM I-80 data, and driving characteristics of the target-vehicle is estimated by using the data. As described in the section \ref{subsection:Proposition of System framework}, the sensing data $z_t$ consists of $m$ number of neighbour vehicles' states at time-step $t$. In this study, seven vehicles' states, $w_t(target)$, $w_t(F_{old})$, $w_t(F_{newL})$, $w_t(F_{newR})$, $w_t(P_{old})$, $w_t(P_{newL})$, and $w_t(P_{newR})$, are considered, therefore, $z_t \in R^{21}$. 
Also, driving characteristics of the target-vehicle $\hat{z}_t(target) \in R^4$ is implemented. 
Three seconds of sensing data $Z_t$ and driving characteristics $\hat{Z}_t$ are fed to the behavior predictor in both training and prediction phases. The dimension of total input data at every time-step is $24\times30$ matrix, where the dimension of $Z$ and $\hat{Z}$ are $21\times30$ matrix and $4\times30$ matrix respectively.

\subsection{Prediction result analysis}
\label{subsection:Prediction result analysis}
For training the driving behavior predictor, 75$\%$ of total data is applied, and 25$\%$ data is used for testing the prediction performance. The trained predictor returns three probabilities, $\hat{y}_t^{LCL}$, $\hat{y}_t^{LCR}$, and $\hat{y}_t^{LK}$ at every time-step, and the final prediction result $\Gamma_t$ is obtained by equation (\ref{eq:prediction}).
\begin{equation}\label{eq:prediction}
\begin{split}
\Gamma_t = \underset{k}{argmax}\left(\hat{y}_t^k\right), k=\left\{LCL, LCR, LK\right\}
\end{split}
\end{equation} 
 Driving behaviors of 407 target-vehicles are predicted, and the prediction quality is evaluated through the confusion matrix \cite{powers2011evaluation}. For lane-change behaviors, LCL, LCR, and LK, the prediction accuracies are [0.9533, 0.9803, 0.9631], the precision-values are [0.7941, 0.8750, 0.9845], the recall values are [0.9153, 0.7000, 0.9695], and the F1-values are [0.8504, 0.7778, 0.9770] respectively.

\subsection{Effectiveness of the driving characteristic}

\begin{figure}[h!]
\centering
\includegraphics[width=7.5cm]{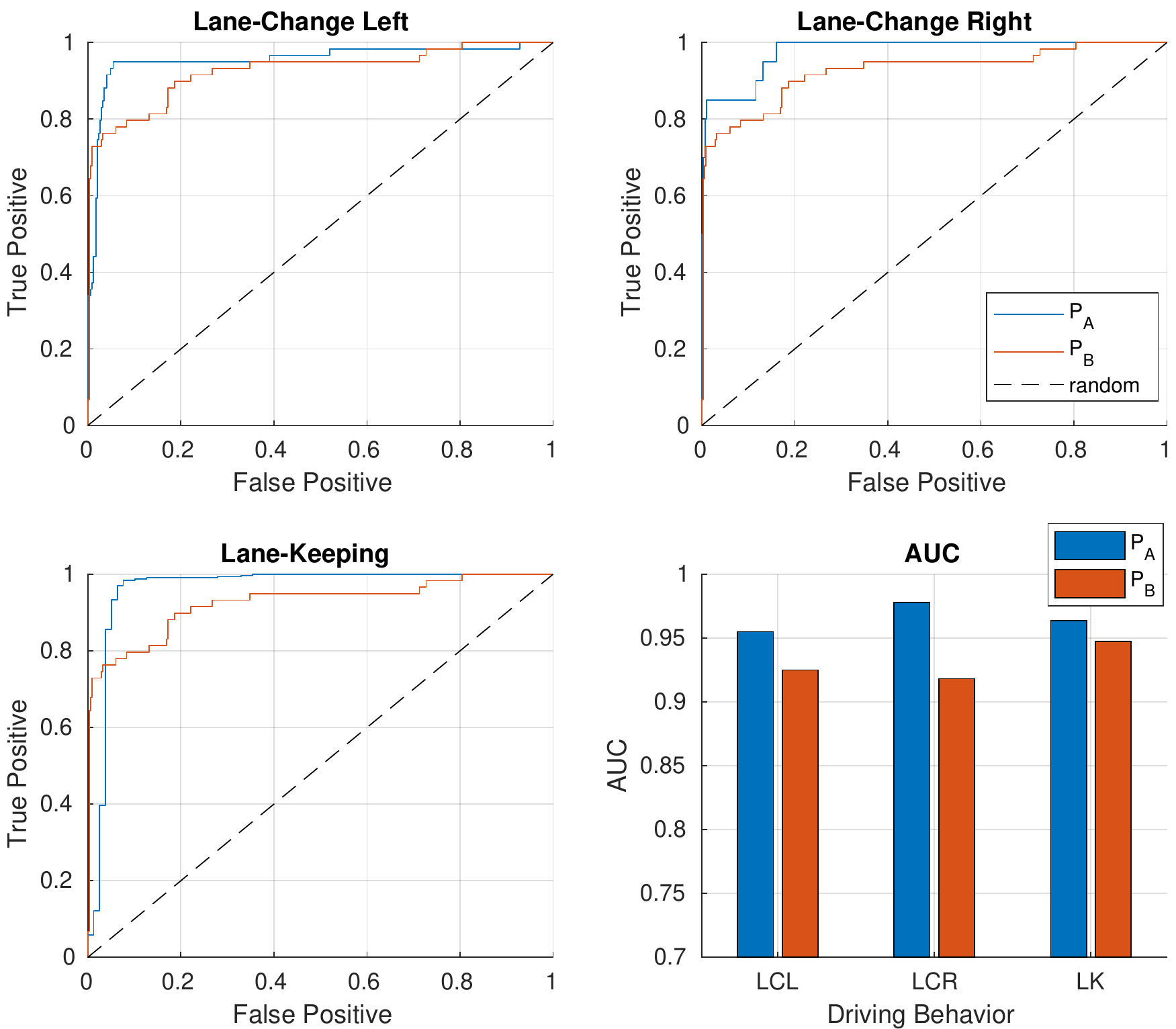}
\caption{ROC curve and AUC of $P_A$ and $P_B$ for LCL, LCR, and LK behaviors.}
\label{fig:roc and auc}
\vspace{-0.4cm}
\end{figure}

To show the worthiness of driving characteristic implementation, the prediction performance of two behavior predictors, $P_A$ and $P_B$, are compared, where $P_A$ is trained and tested by using driving characteristics and sensing data as discussed in Section \ref{subsection:Prediction result analysis} whereas only sensing data is applied for $P_B$. Except the data which is used for training and prediction, everything is identically set for the both predictors. The confusion matrix of $P_B$ shows that the prediction accuracies are [0.8354, 0.9435, 0.8426], the precision-values are [0.4626, 0.4000, 0.9755], the recall values are [0.8305, 0.3000, 0.8206], and the F1-values are [0.8354, 0.9435, 0.8624] respectively for LCL, LCR, and LK. The Receiver Operating Characteristic (ROC) curve and Area Under Curve (AUC) are implemented to present the expected classification performance of $P_A$ and $P_B$ as shown in figure \ref{fig:roc and auc}. The $P_A$'s AUC of LCL, LCR, and LK are [0.9550, 0.9778, 09638], and $P_B$'s are [0.9250, 0.9182, 0.9474] respectively.

As shown by the results, the prediction accuracy of each driving behavior is higher when implementing the driving characteristics with sensing data. Also, the expected classification performance of $P_A$ is better than $P_B$, implying that the implementation of driving characteristic estimation provides more clear and distinct features than only using externally sensable data for classification.    








\section{Conclusion} \label{section:conclusion}

In this study, a system framework is proposed for recognizing and predicting a driver's lane change intentions on the highway. An optimization based method with clustering technique is developed to efficiently estimate the driver characteristic parameters by modeling the driving data, and a Neural Network with LSTM layer is trained to utilize the parameterized driving characteristics for predicting driver behavior transitions. Validation results by real-world traffic data indicates that the designed system is capable of recognizing driving behavior transition patterns and predicting with high accuracy. Furthermore, the comparison study with a Neural Network directly processing unparameterized traffic data demonstrates the prediction advantage of Neural Network using the extracted information from the driver characteristic estimator. 
Overall, the proposed system is useful for assisting an intelligent vehicle's decision making process in naturalistic driving with its high-quality predictions. 




\section*{Acknowledgment}
The work is funded by the National Science Foundation (NSF) Cyber-Physical Systems (CPS)
project under contract \#60046665. The authors would like to thank for the support. 

\bibliography{Driver_jrnl}
\bibliographystyle{ieeetr}

\end{document}